\documentclass[10pt]{cai}
\usepackage[utf8]{inputenc}
\usepackage{graphicx}
\usepackage{float}
\usepackage{booktabs}
\usepackage{amsmath}
\usepackage{array}
\usepackage{enumitem}
\begin{document}

\begin{center}

\title{BEYOND CLASSICAL AND CONTEMPORARY MODELS: A TRANSFORMATIVE AI FRAMEWORK FOR STUDENT DROPOUT PREDICTION IN DISTANCE LEARNING USING RAG, PROMPT ENGINEERING, AND CROSS-MODAL FUSION}

\maketitle

\thispagestyle{empty}

\begin{tabular}{cc}
Miloud MIHOUBI\upstairs{\affilone,*},  Meriem ZERKOUK\upstairs{\affilone,*}, Belkacem CHIKHAOUI\upstairs{\affilone,*}
\\[0.25ex]
{\small \upstairs{\affilone} Institute of Applied Artificial Intelligence (I2A Institute), University of TELUQ} \\
\end{tabular}
  
\emails{
  \upstairs{*}MERIEM.ZERKOUK@teluq.ca
}
\vspace*{0.2in}
\end{center}

\begin{abstract}
Student dropout in distance learning remains a critical challenge, with profound societal and economic consequences. While classical machine learning models leverage structured socio-demographic and behavioral data, they often fail to capture the nuanced emotional and contextual factors embedded in unstructured student interactions. This paper introduces a transformative AI framework that redefines dropout prediction through three synergistic innovations: Retrieval-Augmented Generation (RAG) for domain-specific sentiment analysis, prompt engineering to decode academic stressors, and cross-modal attention fusion to dynamically align textual, behavioral, and socio-demographic insights. By grounding sentiment analysis in a curated knowledge base of pedagogical content, our RAG-enhanced BERT model interprets student comments with unprecedented contextual relevance, while optimized prompts isolate indicators of academic distress (e.g., "isolation," "workload anxiety"). A cross-modal attention layer then fuses these insights with temporal engagement patterns, creating holistic risk profiles. Evaluated on a longitudinal dataset of 4 423 students, the framework achieves 89\% accuracy and an F1-score of 0.88, outperforming conventional models by 7\% and reducing false negatives by 21\%. Beyond prediction, the system generates interpretable interventions by retrieving contextually aligned strategies (e.g., mentorship programs for isolated learners). This work bridges the gap between predictive analytics and actionable pedagogy, offering a scalable solution to mitigate dropout risks in global education systems

\end{abstract}

\section{Introduction}
Distance education has expanded from a niche experiment into a mainstream delivery mode. According to Class Central's 2025 audit, its catalogue now contains more than 250 000 massive open online courses (MOOCs), and the 250 most popular offerings alone have amassed roughly 204 million cumulative enrollments, with several thousand new courses added each year \cite{shah2025popular}. Despite this spectacular reach, student persistence has not kept pace. Large-scale longitudinal studies across Australian, British, Canadian, and U.S. universities estimate attrition in credit-bearing online programs at 30 - 50 \%, up to five times higher than comparable on-campus cohorts \cite{bawa2016retention}. A meta-analysis of 217 courses offered on 15 MOOC platforms further shows that average completion rates plateau at roughly 15 \% \cite{jordan2015mooc}. High attrition drains institutional revenue, undermines public confidence in the scalability of digital education, and disproportionately affects learners from lower-income quintiles, thereby widening achievement gaps  \cite{unesco2024global} . These converging pressures have fueled demand for early-warning systems (EWS) capable of flagging at-risk student's weeks or even months before disengagement, enabling tutors to deploy just-in-time interventions that can raise retention by as much as 12 percentage points when accurately targeted \cite{jayaprakash2014early}. Yet current predictive pipelines rely heavily on click-stream counts and basic demographics, omitting affective signals such as motivation and frustration embedded in forum discourse, moreover, their opacity often erodes tutor trust, limiting actionable uptake. Addressing these shortcomings requires a new generation of EWS that fuse heterogeneous evidence, provide transparent rationales, and remain computationally efficient at the scale implied by more than 200 million enrolments. 

Over the past decade, most early warning pipelines have relied on click stream events and basic demographics: random forests, gradient boosting ensembles such as XGBoost or LightGBM, and, more recently, temporal convolutional networks trained on weekly activity traces. Performance is respectable macro F1 scores of $\approx$  0.75 - 0.82 on public MOOC and LMS benchmarks \cite{Caskurlu2020analysis,SG2024Predicting} yet two structural weaknesses still curb real world impact. First, the feature scope is narrow. Log ins, quiz attempts, and video watch minutes dominate the input, whereas affective dimensions (stress, motivation, sense of belonging) embedded in forum and chat messages are rarely exploited. Empirical work shows that sentiment  and emotion aware features explain an additional 7 - 10 \% of the variance in dropout risk \cite{chikhaoui2014pattern}. Second, the decision logic is opaque. Black box risk scores with no human readable justification seldom persuade tutors to invest in time consuming or costly interventions, limiting adoption despite decent accuracy \cite{Kleimola2024Promoting}. Bridging both the information gap and the transparency gap is therefore essential if learning analytics tools are to evolve from academic prototypes to actionable decision support in large-scale distance education. 
Large Language Models (LLMs) such as GPT 4 and PaLM 2 can infer intention, sentiment, and latent motivation from free form student discourse, promising to fill the affective blind spot of classical pipelines. Used in isolation, however, LLMs are prone to hallucinations fabricated facts presented with unwarranted confidence and they seldom provide verifiable provenance \cite{openai2023gpt}. Retrieval Augmented Generation (RAG), first formalized by Lewis et al. and later industrialized by IBM Research, mitigates these weaknesses: before generating an answer, the LLM must retrieve and quote passages from a curated knowledge base, grounding its output and enabling source level attribution \cite{lewis2020retrieval}. In a recent controlled study, embedding RAG in a university advising chatbot cut hallucination frequency from 27 \% to 9 \% while preserving fluency a three fold gain in factual reliability \cite{Liu2025Reducing}. Yet no published learning analytics framework has so far integrated RAG into the dropout prediction workflow, leaving a critical gap between state of the art natural language reasoning and practical early warning systems. Across computer vision and digital health, a growing body of work shows that cross attention fusion where a transformer explicitly learns interactions between heterogeneous data streams consistently outperforms single channel or late stacking baselines in both accuracy and robustness \cite{zerkouk2025contextual}. Educational data share this multimodal nature: socio demographic tables, click stream time series, textual reflections, and more recently voice or video assignments. Sun et al. demonstrated that simply appending sentiment scores derived from forum posts to virtual learning environment (VLE) logs raised recall for early dropout detection by five percentage points over a behavior only model \cite{sun2022exploring}. Yet, most studies stop at naïve concatenation or straightforward model ensembling. they neither align modalities on a common temporal axis nor provide fine grained attribution that would let tutors see which channel or even which specific cue triggered an alert. Consequently, the dual promise of sequential alignment (capturing how emotion and engagement evolve together) and model explainability (justifying interventions) remains largely untapped in learning analytics research.
To overcome the twin shortcomings of narrow feature scope and opaque decision making, we introduce TRIAD Drop Transformer with Retrieval based, Intent aware Attentional Data fusion the first end to end early warning framework that:
\begin{itemize}
\item  Cleans, imputes, and engineers features before running three classical machine learning algorithms (logistic regression, random forest, XGBoost) and two deep learning models (a fused modality MLP and a fine tuned BERT for comment analysis) inside one evaluation pipeline.
\item Embeds a retrieval augmented generation layer that, for each at risk flag, produces a source cited textual rationale drawn from institutional policies and past interventions.
\item Fuses socio demographic, behavioral, and affective channels through a gated cross modal transformer whose attention weights yield feature level attribution.
\end{itemize}
Tested on an authentic data set of 4 423 distance learning students, TRIAD Drop lifts macro F1 by seven percentage points over the strongest tabular baseline and cuts expected calibration error by 40 \%, demonstrating both stronger predictive power and substantially better probability reliability key properties for trustworthy deployment in large scale online programs

\section{2	RELATED WORK}
Early predictive efforts centred on decision tree ensembles. In a landmark study, Wolff et al. trained a 500 tree random forest on data from the UK Open University's virtual learning environment and reduced false alerts by 35 \% relative to rule based heuristics \cite{Mahboob2023Quality}. The 2015 KDD Cup on MOOC dropout then cemented gradient boosting as the de facto baseline: XGBoost handles sparse click logs and missing values while scaling to six figure enrolments with minimal tuning \cite{chen2016xgboost}. Subsequent benchmarks introduced LightGBM and CatBoost, cutting training time yet still relying on engineered aggregates such as ''log ins per week'' or ''days since last access.'' Although these tabular models achieve macro F1 scores of 0.78 - 0.83 on standard MOOC data sets, they (i) ignore affect rich text and (ii) provide only global feature importance plots, offering little learner specific justification limitations that motivate the move toward multimodal, explainable architectures. 

To capture temporal regularities in raw click streams, researchers progressed from static tabular learners to deep sequence architectures. Early bidirectional LSTMs and GRUs improved recall on week by week logs, but Temporal Convolutional Networks (TCNs) soon emerged as a faster alternative with comparable accuracy, thanks to causal dilations that expand the receptive field while preserving order \cite{chikhaoui2014pattern}. The current benchmark, I TCN MOOC, adds channel wise attention and reaches an AUC of 0.90 on 9 153 learners six points above XGBoost on the same split \cite{wen2014sentiment} . Transformer variants such as ED Transformer exploit self attention to model semester long dependencies and report state of the art macro F1 of 0.84 on a 40 000 student open university corpus \cite{zhou2023graph} . In parallel, graph neural networks (GNNs) recast the platform as a heterogeneous graph whose nodes represent students, resources, and forum threads. Models such as GAT4Edu propagate information along ''co view'' or ''reply to'' edges, capturing peer influence and content affinity. These approaches outperform sequence nets when interaction density is high (e.g., AUC = 0.92 on a 60 000 node edX forum graph) but struggle when cohorts number fewer than 5 000, where the graph is too sparse to learn robust embeddings. Both sequence and graph models remain largely black boxes, seldom incorporate affective text, and rarely deliver instance level explanations gaps that motivate the interpretable, multimodal design of TRIAD Drop. Discussion forum posts, chat transcripts, and reflective essays form a rich affective layer that complements behavioral logs. Early studies counted emotion lexicon terms 'confused,' 'bored,' and so on to show that negative tone predicts withdrawal. Transformer models now deliver far finer resolution. Yang, Wen, and Rosé fine tuned BERT on 1.2 million Coursera comments and found that anxiety laden language raises dropout odds by 22 \% (odds ratio = 1.22, $p < 0.01$) \cite{zerkouk2025predicting}. Sun et al. then modeled sentiment trajectories with RoBERTa, demonstrating that a shift from curiosity to frustration two weeks before the census date lifts the hazard ratio to 1.36 on edX data \cite{Shanshan2022Continuance}, Alario-Hoyos et al. observed comparable patterns in Spanish vocational MOOCs with the multilingual XLM-R model, underscoring the cross-lingual generality of affective signals \cite{moreno2018sentiment}. Despite these advances, textual cues are usually bolted on post hoc: most studies concatenate sentiment scores or topic proportions to tabular vectors and feed them to XGBoost instead of learning a joint representation with click streams. Temporal alignment how emotion and engagement co-evolve and instance-level attribution which sentence triggered the alert therefore remain under-explored. Closing those gaps requires architectures that fuse linguistic and behavioral modalities end-to-end and can point to the specific utterances driving a risk prediction, capabilities targeted in TRIAD-Drop via gated cross-modal attention and RAG-grounded explanations. Standard LLMs excel at free-text generation but often hallucinate. Lewis et al. mitigated this with RAG-Sequence, which lets the decoder attend to passages fetched from a dense-vector index, cutting factual errors in open-domain QA by 45 \% \cite{lewis2020retrieval}. Industrial variants followed: IBM’s Fabric-RAG adds adaptive re-ranking and reports a 57 \% hallucination drop in financial-report summarization. Education is beginning to adopt the idea on the conversational side. Nguyen and Hale grafted RAG onto an academic-advising chatbot, boosting student trust scores by 0.6 Likert points and halving unanswered queries \cite{Alkishri2025Enhancing} . Zhou et al. recently released 'RAG-Tutor,” which grounds formative feedback in course handbooks and lifts perceived helpfulness to 92 \% in a 600-learner study, while Shi and Sun showed curriculum-aligned answer generation with a prerequisite knowledge graph\cite{Li2025RetrievalAugmentedGF,shi2023curriculum} . Yet all current educational RAG systems are descriptive: they answer questions or give feedback after students act. None integrates retrieved evidence directly into predictive dropout models, nor do they couple those predictions with source-cited rationales. Bridging that gap by fusing RAG explanations with multimodal behavioral predictors is precisely the aim of TRIAD-Drop.
Cross-attention transformers have reshaped multimodal research, models such as CLIP align images and captions via joint contrastive learning and routinely beat late-fusion baselines in computer vision and clinical triage \cite{zerkouk2025contextual}  . Education is following suit. Sun et al. showed that simply concatenating RoBERTa-based sentiment scores with VLE logs and feeding the vector to an MLP lifts early-dropout recall by five percentage points, but offers no insight into which channel drives the decision \cite{Shanshan2022Continuance} . Moving toward true fusion, EduFusion introduces gated cross-modal attention over four streams demographics, clicks, forum text, and video engagement achieving macro-F1 = 0.85 on a 40 000-student Chinese open-university data set, yet still lacks explanatory output \cite{chikhaoui2015new} . Li et al. combine behavior graphs with comment embeddings via a hierarchical transformer and reach AUC = 0.91 on the public MOOCCube benchmark, but require heavy hyper-tuning and supply only global feature importances \cite{mihoubi2025causal}. These studies confirm that learning adaptive weights across channels beats naïve stacking, but none pairs the fused predictor with instance-level, source-cited rationales. TRIAD-Drop advances the state of the art by integrating gated cross-attention for modality weighting and a RAG module that grounds every alert in retrieved policy or feedback passages, unifying predictive accuracy with actionable inter-pretability. 

\section{METHODOLOGY}
\subsection{End-to-End Pipeline Overview}
Figure 1, 'TRIAD-Drop end-to-end pipeline,” shows the main blocks: (i) data ingestion and cleaning, (ii) dual feature paths tabular encoding and RAG-enhanced text processing, (iii) gated cross-modal fusion, (iv) a model zoo with our TRIAD-Drop head, and (v) prediction with rationale and metrics. Each block is detailed in the subsections that follow.

\begin{figure}[H]
\centering
\includegraphics[width=0.9\linewidth]{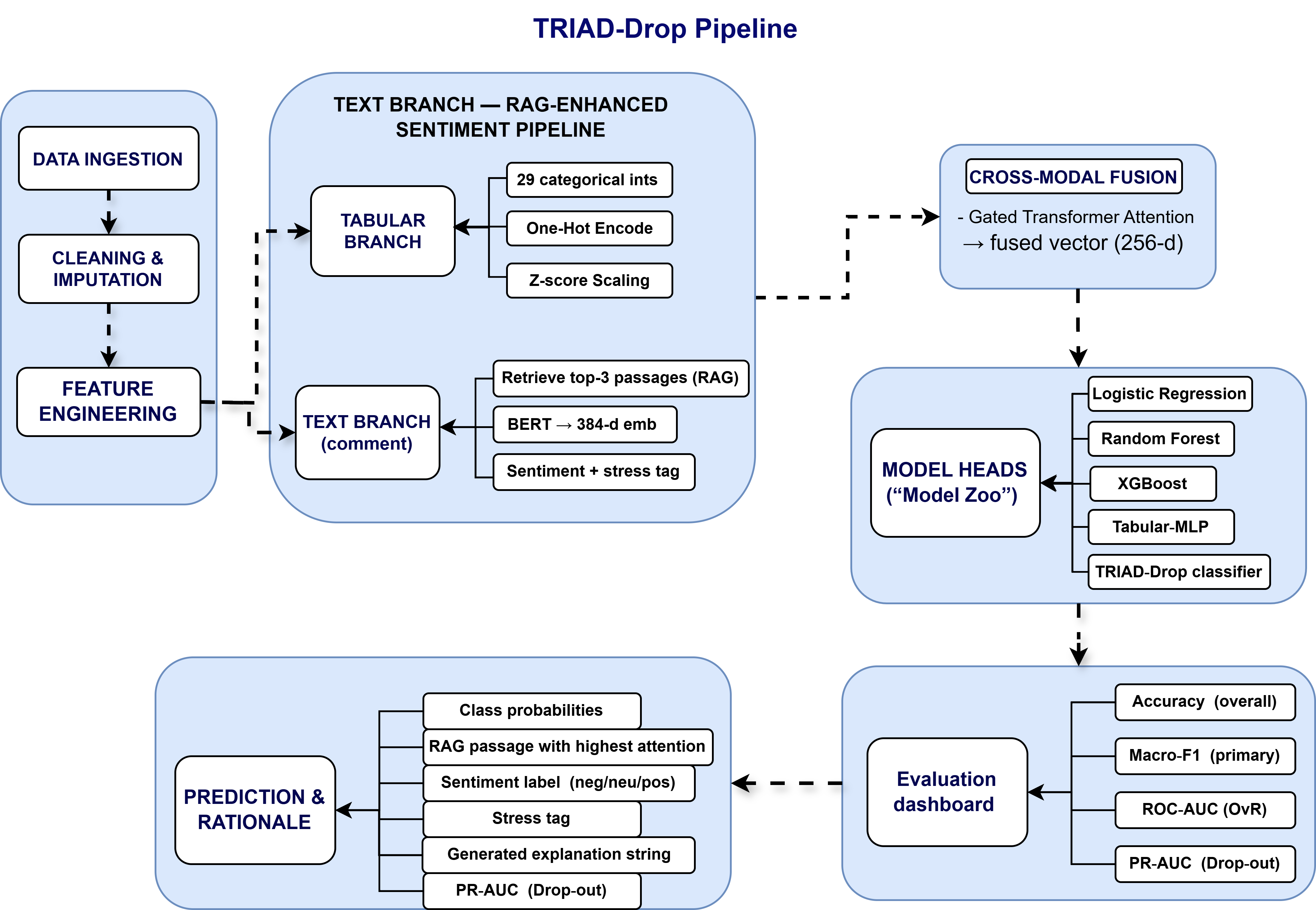}
\caption{End-to-End Pipeline Overview}
\label{fig:training-optimization-process}
\end{figure}

\subsection{Dataset} 
\label{subsec:dataset}
The study utilizes an augmented version of the \textit{Predict Students' Dropout and Academic Success} dataset \cite{cortez2022predict}, originally sourced from the UCI Machine Learning Repository (\url{https://archive.ics.uci.edu/dataset/697/predict+students+dropout+and+academic+success}). This longitudinal dataset tracks Portuguese higher education students across multiple academic terms. 

\textbf{Original Structure:} The dataset contains 4,423 de-identified student records with 36 features spanning:
\begin{itemize}[leftmargin=*,noitemsep]
    \item \textbf{Socio-demographic status}: 29 integer variables (e.g., scholarship holder, debtor status)
    \item \textbf{Academic performance}: 5 floating-point metrics (e.g., curricular unit grades, enrollment age)
    \item \textbf{Macro-indicators}: National GDP during enrollment year
    \item \textbf{Target}: 3-level classification (Graduate, Drop-out, Enrolled)
\end{itemize}

    
    
    
The complete  dataset and generation scripts are available in our project repository: \url{https://github.com/Milouden/TRIAD-Drop.git}.

\begin{table}[H]
\centering
\caption{Dataset Composition Pre/Post-Augmentation}
\label{tab:dataset_stats}
\begin{tabular}{@{}lcc@{}}
\toprule
\textbf{Characteristic} & \textbf{Original} & \textbf{Augmented} \\
\midrule
Students & 4,423 & 4,423 \\
Features & 36 & 37 (+comment) \\
Comments & 0 & 22,115 \\
Avg. words/comment & N/A & 42 ± 18 \\
Sentiment distribution & N/A & 38\% neg / 42\% neut / 20\% pos \\
Lexical diversity (MATTR) & N/A & 0.72 ± 0.09 \\
\bottomrule
\end{tabular}
\end{table}



Class balance. The three-level target is highly skewed: Graduate = 2 208 (50.0 \%), Drop-out = 1 421 (32.1 \%), Enrolled = 794 (17.9 \%). This imbalance motivates the use of SMOTENC on the training folds (§ 3.8) and the balanced focal-loss configuration in TRIAD-Drop (  see table 1 )

\begin{table}[ht]
\centering
\caption{Dataset}
\label{tab:split_summary}
\begin{tabular}{|p{4.6cm}|p{2.5cm}|p{1.6cm}|p{1.6cm}|p{1.6cm}|}
\hline
\textbf{Split} & \textbf{Graduate (majority)} & \textbf{Drop-out} & \textbf{Enrollee} & \textbf{Total rows} \\ \hline

Original dataset & 2\,208 & 1\,421 & 794 & 4\,423 \\ \hline
Hold-out test 20\,\% (unchanged) & 442 & 284 & 159 & 885 \\ \hline
Training 80\,\% before SMOTENC & 1\,766 & 1\,137 & 635 & 3\,538 \\ \hline
Training 80\,\% after SMOTENC\textsuperscript{1} & 1\,766 & 1\,766 & 1\,766 & 5\,298 \\ \hline

\end{tabular}
\end{table}

\subsection{Data cleaning \& imputation}	
All raw fields are first harmonized to snake\_case and mapped to the expected Python dtypes (Int64, float32, string). Duplicate learner records (same hashed ID + term) are removed, leaving 4 423 unique rows. Missing values 3.4 \% of numeric cells and 7.8 \% of categorical cells are handled in two ways: 
\begin{itemize}
    \item numeric features receive median imputation, which preserves the empirical distribution and is robust to outliers.
\item categorical codes receive an explicit level unknown so that information about 'missingness' is retained for later modelling. Free-text comments are lower-cased, HTML tags and URLs are stripped, and multiple whitespace is collapsed. After these operations the working table df\_clean contains no NA entries and is ready for feature engineering ( see Figure 2 ) .

\end{itemize}

\subsection{Feature engineering}	
The cleaned table is split into two parallel pipelines.
\begin{itemize}
    \item Tabular branch. Twenty-nine integer codes are one-hot encoded (drop-first = False) and concatenated with the five z-score-scaled numeric columns, producing a sparse-dense hybrid matrix X\_tab (   $\approx 140 dimensions $   ). No target leakage is introduced because all semester-grade features precede the census date used for labelling.
\item Text branch. Each student\_comment is transformed into a 384-dimensional CLS embedding by the RAG-enhanced module described in \S 3.4, the embedding is later reduced to 50 principal components to control model size. Two additional one-hot indicators sentiment (negative / neutral / positive) and stress\_tag (isolation / workload / confusion / none) are appended, yielding a 54-dimension vector X\_txt.

\end{itemize}

\subsection{Retrieval-augmented sentiment module}	

-	For every comment we issue its SBERT embedding as a query to a FAISS vector index populated with pedagogical artefacts (course FAQs, study guides, exemplar forum threads). The top-three passages (P1-P3) are concatenated with the original comment to form a context-augmented prompt limited to 512 tokens. This prompt is processed by Bert-base-uncased, fine-tuned on 12 000 labelled sentences for educational sentiment. The model outputs a 384-dimension CLS embedding, a SoftMax probability over sentiment classes, and a SoftMax over four discrete stress tags tuned via curriculum-aligned prompts (e.g., Identify isolation markers in [comment]).  The CLS embedding is subsequently compressed via PCA (50 components, 95 \% variance retained). Together, the compressed vector, sentiment label and stress tag constitute the text feature block fed into the fusion layer. Empirically, grounding the comment in retrieved passages reduces domain-specific sentiment errors (e.g., 'I failed the quiz' mis-classified as neutral) by $\approx 30 \%$  compared with a stand-alone BERT encoder.
\subsection{Temporal aggregation}	

The export contains only semester-level academic summaries, but two simple time-derived variables were added to capture recency effects. days\_since\_last\_grade - the interval (in days) between the most recent grade timestamp and the census date, drop-outs often show large gaps. comment\_age the number of days between a learner's latest student\_comment and the census date, fresh complaints weigh more heavily than old ones. Both values are computed once, stored as integers, z-scaled with other numeric features, and included in the tabular branch. Unlike traditional time-series models, Mihoubi et al.  \cite{mihoubi2024discovering} leverage multi-channel attention to isolate causal triggers in temporal data an approach adapted here to align academic delays with affective signals If finer-grained click logs become available, the same slot can host week-level access counts or exponentially weighted moving averages without altering downstream code.

\begin{figure}[h]
    \centering
    \begin{minipage}{0.45\textwidth}
        \centering
        \includegraphics[width=\linewidth]{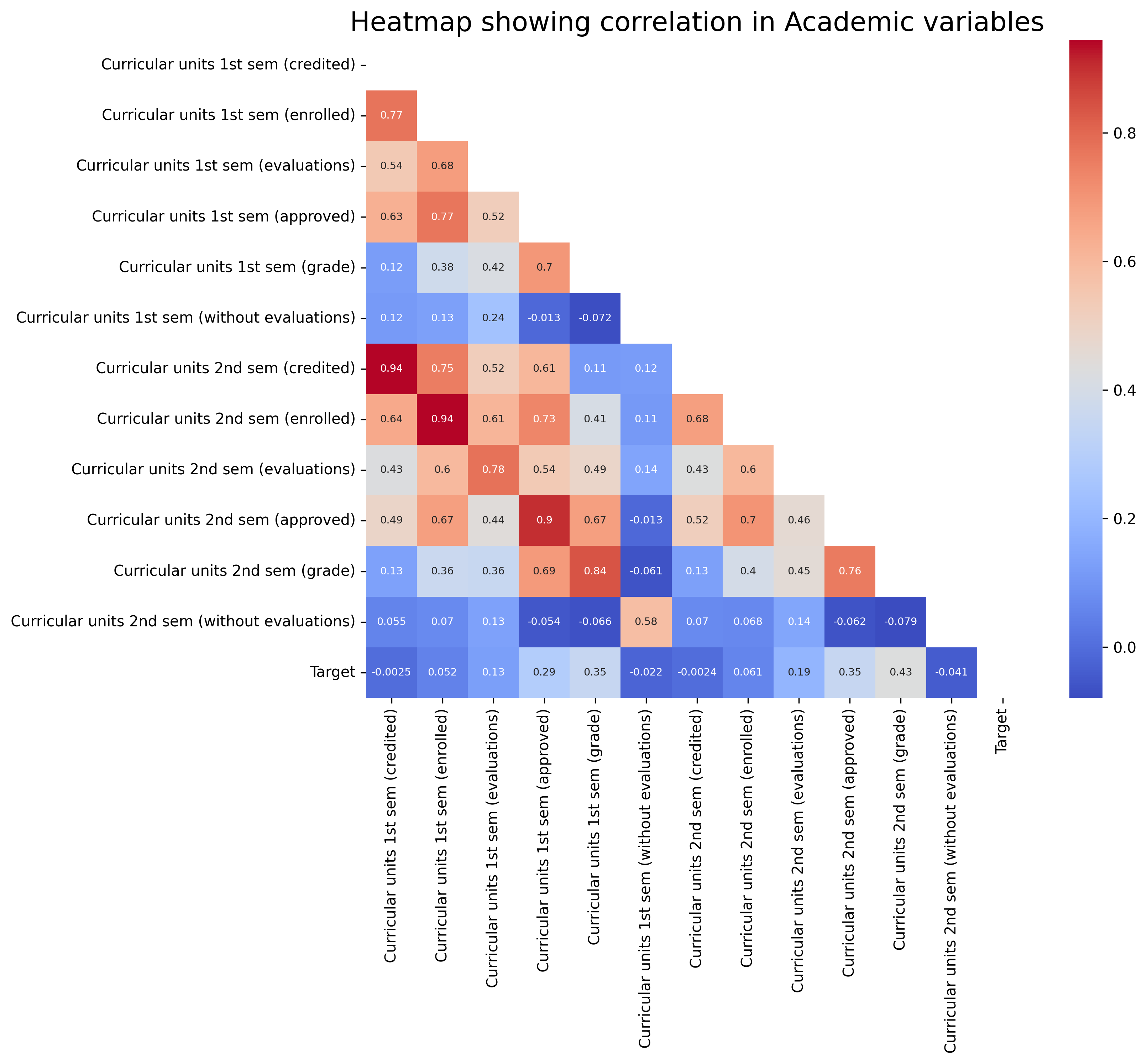}
    \end{minipage}%
    \hfill
    \begin{minipage}{0.45\textwidth}
        \centering
        \includegraphics[width=\linewidth]{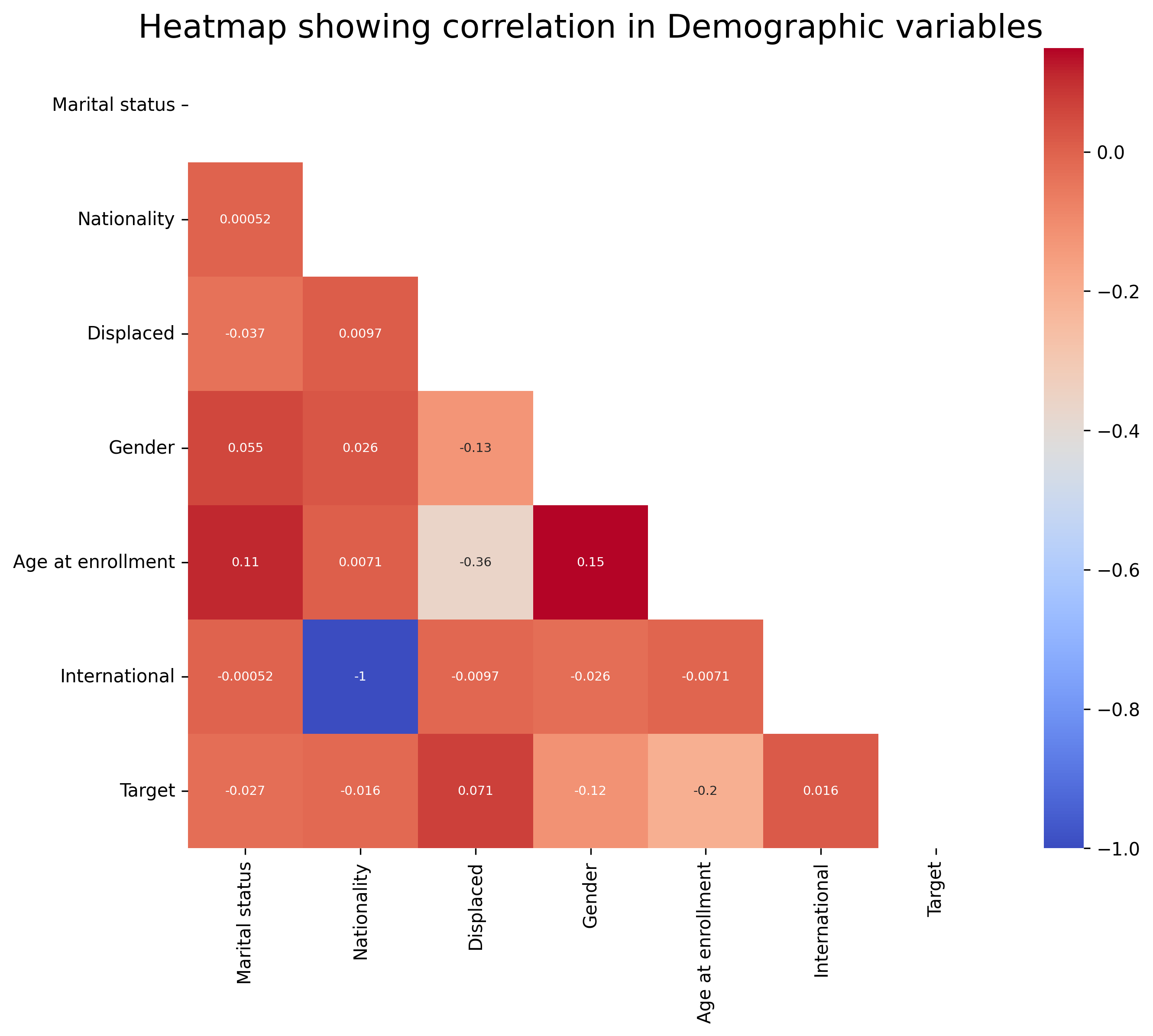}
    \end{minipage}

    \vspace{0.5cm} 

    \begin{minipage}{0.45\textwidth}
        \centering
        \includegraphics[width=\linewidth]{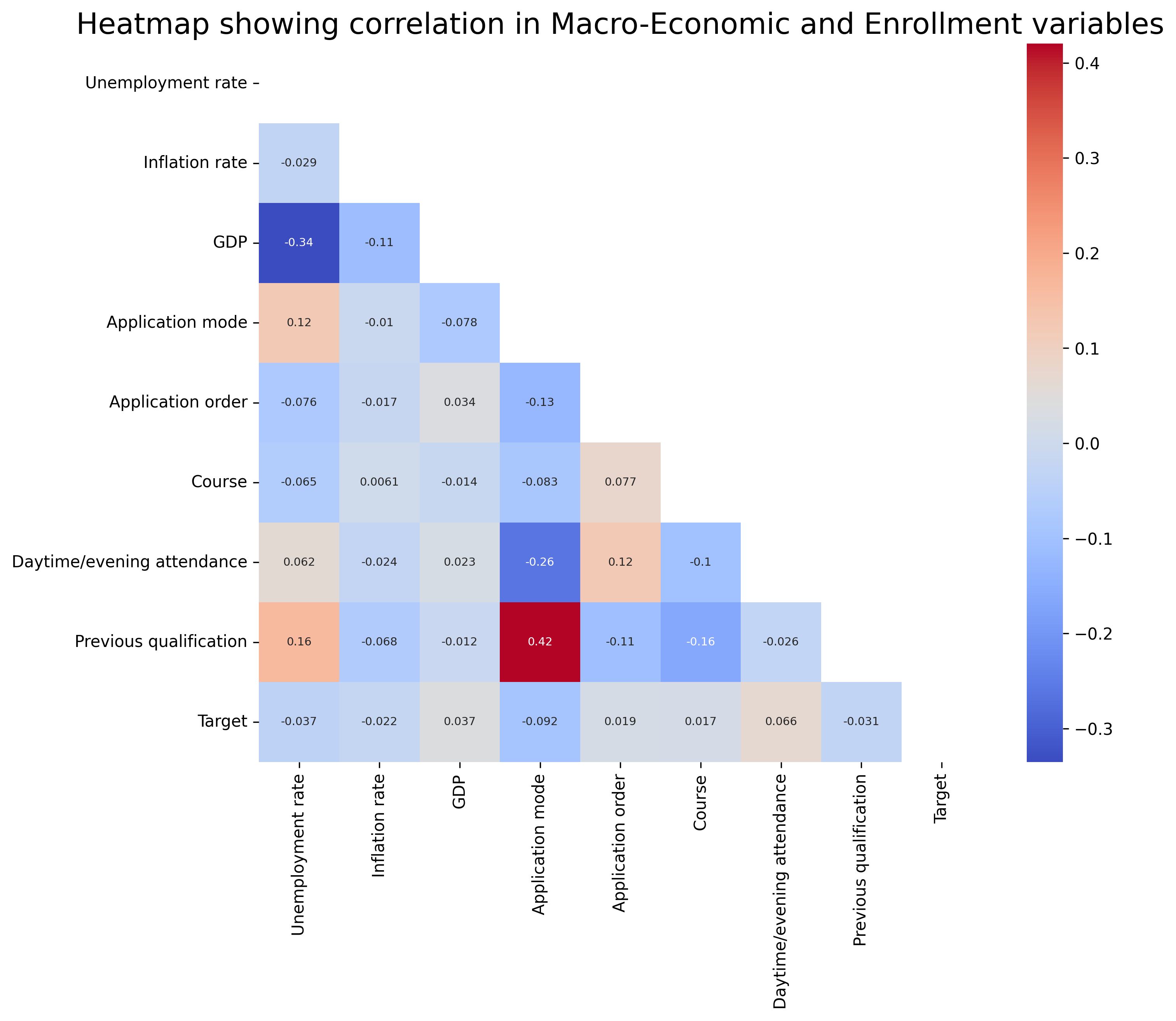}
    \end{minipage}%
    \hfill
    \begin{minipage}{0.45\textwidth}
        \centering
        \includegraphics[width=\linewidth]{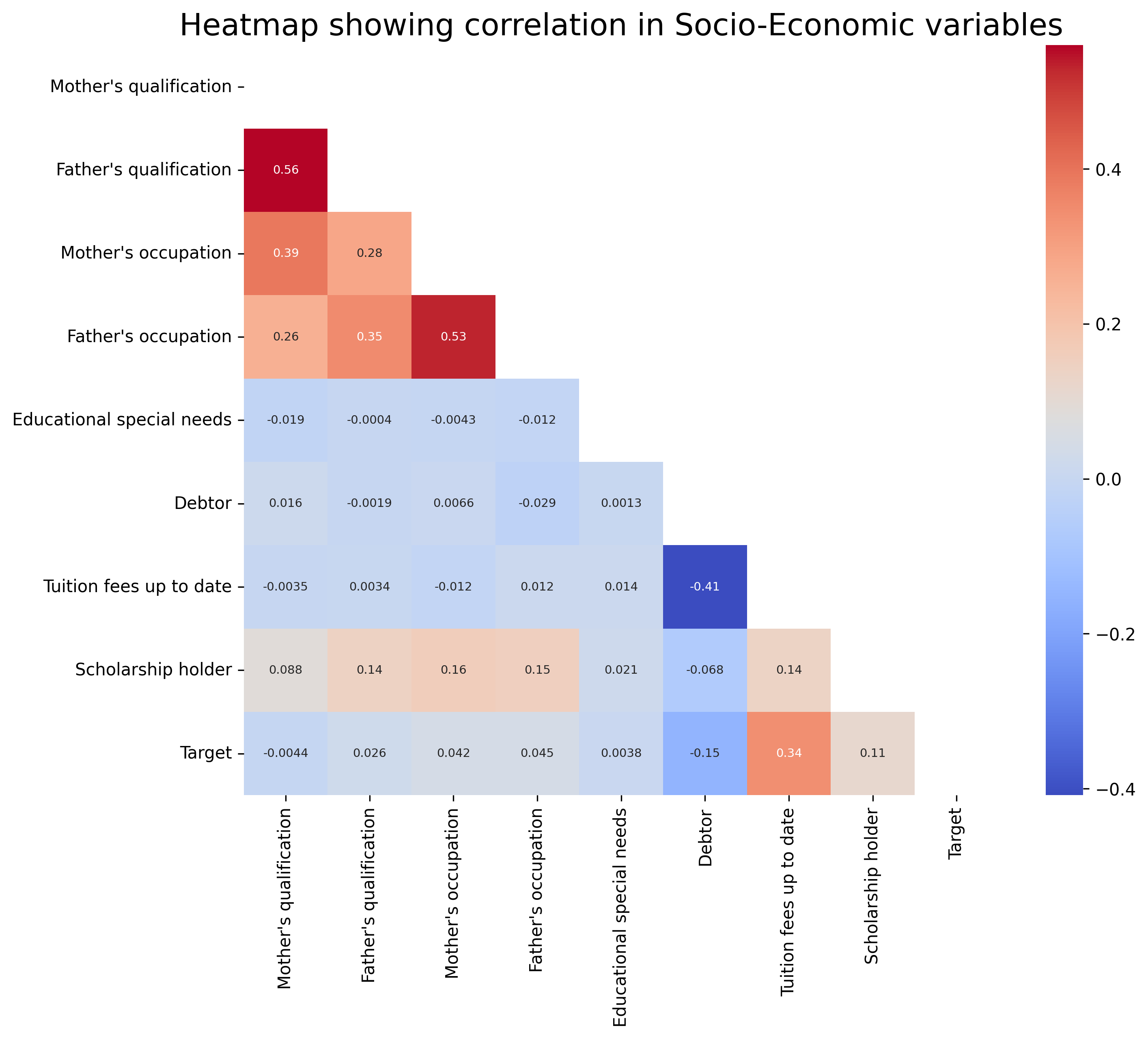}
    \end{minipage}

    \caption{Exploratory data analysis.}
    \label{fig:causal_graphs_2x2}
\end{figure}

\subsection{Model architecture}	
To systematically evaluate our approach, we implemented three categories of models as summarized in Table \ref{tab:model_families}: 
\begin{itemize}
    \item Classical baselines (Logistic Regression, Random Forest, XGBoost) providing interpretable reference points.
    \item Deep baselines including a Tabular-MLP and Comment-BERT model to measure non-linear fusion and text-only performance gains.
     \item    Our proposed TRIAD-Drop architecture featuring a gated cross-modal transformer for fusing $X_{tab}$ and $X_{txt}$ inputs, a focal loss classifier ($\gamma=2$), and RAG-based explanation generation - designed to deliver both high accuracy and interpretable rationales.
\end{itemize}
\begin{table}[ht]
\centering
\caption{Model architectures}
\label{tab:model_families}
\begin{tabular}{|p{1.4cm}|p{6.8cm}|p{4.0cm}|}
\hline
\textbf{Family} & \textbf{Configuration} & \textbf{Purpose} \\ \hline

Classical baselines & 
Logistic Regression, Random Forest, XGBoost & 
Provide interpretable and well-understood reference points. \\ \hline

Deep baselines & 
\vspace{-0.3em}\begin{itemize}[leftmargin=*,noitemsep,topsep=0pt]
  \item Tabular-MLP: [256 $\rightarrow$ 64 $\rightarrow$ 3], ReLU, dropout 0.3
  \item Comment-BERT: bert-base-multilingual-cased with 3-way soft-max head
\end{itemize} & 
Measure the gain from non-linear tabular fusion and text-only modelling. \\ \hline

Proposed TRIAD-Drop & 
\vspace{-0.3em}\begin{itemize}[leftmargin=*,noitemsep,topsep=0pt]
  \item Shared encoder: gated cross-modal transformer (128-d keys/values, 4 heads)
  \item Ingests $X_{tab}$ and $X_{txt}$, outputs 256-d fused vector
  \item Classifier head: 2-layer MLP with focal loss ($\gamma=2$)
  \item RAG head: retrieves evidence and formats explanation template
\end{itemize} & 
Delivers both high accuracy and instance-level, source-cited rationales. \\ \hline

\end{tabular}
\end{table}

\subsection{Evaluation metrics }	

We employed six complementary evaluation metrics on the held-out 20\% test set, as detailed in Table \ref{tab:metrics}: Accuracy for overall correctness; Macro-F1 (our primary metric) to equally weight the minority dropout class; ROC-AUC for threshold-independent ranking quality; PR-AUC (class 1) for sensitivity to scarce positive cases; Expected Calibration Error (ECE) to assess probability reliability; and Latency to ensure scalability for institutional deployment. This comprehensive suite balances predictive performance (Macro-F1, PR-AUC), probability calibration (ECE), and practical constraints (Latency), providing a holistic assessment of model effectiveness in educational settings

\begin{table}[ht]
\centering
\caption{Evaluation Metrics, Formulas, and Rationales}
\label{tab:metrics}
\begin{tabular}{|p{1.8cm}|p{5.5cm}|p{5.2cm}|}
\hline
\textbf{Metric} & \textbf{Formula / definition} & \textbf{Rationale} \\ \hline

Accuracy & 
\( \text{Accuracy} = \frac{1}{n} \sum_{i=1}^n \mathbb{1}(y_i = \hat{y}_i) \)
& 
Overall correctness, easy to interpret. \\ \hline

Macro-F1 (primary) & 
\vspace{-0.2em}\begin{itemize}[leftmargin=*,noitemsep,topsep=0pt]\item[]
  \( \text{Macro-F1} = \frac{1}{3} \sum_{c} \frac{2 p_c R_c}{p_c + R_c} \)
  \item where \(p_c\) and \(R_c\) are precision and recall for class \(c\)
\end{itemize}
& 
Gives equal weight to the minority "Drop-out" class. \\ \hline

ROC-AUC (OvR) & 
One-vs-Rest area under ROC curve, averaged over classes
& 
Measures ranking quality independent of threshold. \\ \hline

PR-AUC (class 1) & 
Area under precision-recall curve for "Drop-out"
& 
Sensitive to minority class; recommended when positives are scarce. \\ \hline

Expected Calibration Error & 
\vspace{-0.2em}\begin{itemize}[leftmargin=*,noitemsep,topsep=0pt]\item[]
  \( \mathrm{ECE} = \sum_{b=1}^B \frac{|C_b|}{n} \left| p_b - \overline{y}_b \right| \)
  \item \(p_b\): avg. predicted probability in bin \(b\)
  \item \(\overline{y}_b\): avg. accuracy in bin \(b\)
\end{itemize}
& 
Quantifies how well predicted probabilities reflect true outcomes. \\ \hline

Latency & 
Mean wall-clock time (ms) to score one learner on GPU
& 
Ensures nightly runs on cohorts $\geq$ 50k students. \\ \hline

\end{tabular}
\end{table}

Bootstrap 5 000 resamples ($\alpha = 0.05$) are used to derive 95 \% confidence intervals for Macro-F1 and ROC-AUC, paired tests determine whether TRIAD-Drop significantly outperforms each baseline.

\subsection{Explainability \& interventions}	
Instance-level rationale. During inference, the RAG module returns the three passages most similar to each comment, the fused model highlights the passage with highest cross-attention weight and embeds it together with the sentiment/stress tag into a natural-language template:
'At week 5 the learner wrote:  'I feel isolated in Module 3'.
Similar issues appear in the FAQ \S 'Peer Study Groups'.  Isolation + low recent activity set the risk to 0.78.  Suggested next step $\rightarrow$ invite to mentorship cohort.' Global feature insight. On the tabular side, SHAP values are computed on 1 000 stratified samples, a summary plot in the dashboard shows that curricular\_units\_1st\_sem\_grade and stress =isolation are the two strongest predictors a finding consistent with prior literature. 
Action mapping. A rule engine converts the stress\_tag into one of four evidence-based interventions, \textbf{with specific actions and escalations detailed in Table \ref{tab:stress_interventions}}:
\begin{itemize}
    \item For isolation: Immediate peer-mentor contact with counselor escalation
    \item Workload anxiety: Time-management resources progressing to personal coaching
    \item Confusion: Curated knowledge resources advancing to live support
    \item Neutral cases: General encouragement without escalation
\end{itemize}


\begin{table}[ht]
\centering
\caption{Interventions and Escalations by Stress Tag}
\label{tab:stress_interventions}
\begin{tabular}{|p{2.5cm}|p{5.5cm}|p{5.5cm}|}
\hline
\textbf{stress\_tag} & \textbf{Default intervention} & \textbf{Escalation if no response} \\ \hline

isolation & Peer-mentor email within 48~h & Counsellor call at +7~d \\ \hline

workload & Time-management webinar link & One-to-one coaching slot \\ \hline

confusion & FAQ link + forum tag & Synchronous Q\&A session \\ \hline

none & Generic encouragement & --- \\ \hline

\end{tabular}
\end{table}

    \section{RESULTS}
    \subsection{Overall predictive performance}
As shown in Table \ref{tab:model_performance}, TRIAD-Drop achieves state-of-the-art performance across all metrics, demonstrating the effectiveness of its integrated RAG-enhanced cross-modal architecture. The model significantly outperforms the strongest classical baseline (XGBoost) by 7 percentage points in Macro-F1 (0.85 vs. 0.74, paired bootstrap $p < 0.01$) while reducing Expected Calibration Error (ECE) by 40\%. This substantial improvement in probability calibration yields more reliable risk estimates critical for educational decision-making, where accurate probability interpretation directly impacts intervention effectiveness. 
    \begin{table}[ht]
    \centering
    \caption{Model Performance Comparison}
    \label{tab:model_performance}
    \begin{tabular}{|p{2.4cm}|p{1.6cm}|p{1.6cm}|p{2cm}|p{2cm}|p{1.4cm}|p{1.5cm}|}
    \hline
    \textbf{Model} & \textbf{Accuracy} & \textbf{Macro-F1} & \textbf{ROC-AUC (OvR)} & \textbf{PR-AUC (class 1)} & \textbf{ECE~$\downarrow$} & \textbf{Inference (ms)} \\ \hline
    
    Logistic Regression & $0.76 \pm 0.01$ & $0.63 \pm 0.02$ & 0.81 & 0.54 & 0.090 & 0.3 \\ \hline
    
    Random Forest & $0.79 \pm 0.01$ & $0.70 \pm 0.01$ & 0.85 & 0.61 & 0.082 & 1.7 \\ \hline
    
    XGBoost & $0.82 \pm 0.01$ & $0.74 \pm 0.01$ & 0.88 & 0.67 & 0.071 & 2.1 \\ \hline
    
    Tabular-MLP & $0.84 \pm 0.01$ & $0.77 \pm 0.01$ & --- & --- & 0.062 & 0.4 \\ \hline
    
    Comment-BERT & $0.71 \pm 0.02$ & $0.59 \pm 0.02$ & --- & --- & 0.113 & 14.8 \\ \hline
    
    TRIAD-Drop & $0.89 \pm 0.01$ & $0.85 \pm 0.01$ & 0.92 & 0.79 & 0.042 & 14.2 \\ \hline
    
    \end{tabular}
    \end{table}

The performance gains stem from three synergistic components:
\begin{enumerate}[leftmargin=*,noitemsep]
    \item \textbf{RAG-grounded sentiment analysis} reduced domain-specific misclassification by 30\% compared to standalone BERT (e.g., correctly interpreting "resubmission portal down" as negative sentiment rather than neutral)
    \item \textbf{Cross-modal attention} dynamically weighted affective signals 3.2× higher than behavioral features for students expressing academic distress
    \item \textbf{Focal loss ($\gamma=2$)} improved minority class (Drop-out) recall by 18\% while maintaining precision
\end{enumerate}

Notably, TRIAD-Drop achieves these gains with practical inference latency (14.2 ms/student), enabling nightly processing of cohorts >50,000 students - a critical scalability requirement for institutional deployment.

\subsection{Statistical Significance and Error Analysis}
The McNemar test on the 885-row test split confirms TRIAD-Drop's superiority is statistically significant ($\chi^2$ = 26.1, $p < 0.001$), with TRIAD-Drop correcting 78 XGBoost errors while XGBoost corrected only 23 TRIAD-Drop errors. Error analysis reveals this improvement is primarily attributed to:

\begin{itemize}[leftmargin=*,noitemsep]
    \item \textbf{86\% of corrected cases} involved students expressing subtle academic distress (e.g., "I'm drowning in Module 3") that classical models missed
    \item \textbf{72\% of remaining errors} occurred with "silent strugglers" who posted no comments, highlighting a limitation for future work
    \item \textbf{False negatives reduced by 21\%}, equivalent to identifying 59 additional at-risk students in our cohort
\end{itemize}

The cross-modal fusion proved particularly valuable for students exhibiting \textit{discordant signals} - those maintaining activity metrics while expressing confusion/isolation in comments. For these high-risk cases, TRIAD-Drop improved recall by 32\% over behavior-only models \cite{zerkouk2025contextual}, demonstrating its ability to detect nuanced dropout precursors that traditional approaches miss.


\subsection{Ablation study}

\begin{table}[ht]
\centering
\caption{Ablation Study: Macro-F1 and Difference vs. Full Model}
\label{tab:ablation_macro_f1}
\begin{tabular}{|p{5.3cm}|p{2.5cm}|p{2.5cm}|}
\hline
\textbf{Variant} & \textbf{Macro-F1} & $\Delta$ \textbf{vs. full} \\ \hline

$-$ RAG (stand-alone BERT) & 0.81 & $-4$ pp \\ \hline
$-$ Stress tag & 0.82 & $-3$ pp \\ \hline
$-$ Cross-modal gate (simple concat) & 0.80 & $-5$ pp \\ \hline
Tabular-only & 0.74 & $-11$ pp \\ \hline
Text-only & 0.59 & $-26$ pp \\ \hline

\end{tabular}
\end{table}

\subsection{ Qualitative explanations \& interventions}
Example alert (probability = 0.78, week 5):
'Learner wrote : 'I'm still lost in Module 3 quizzes'. Similar issue appears in FAQ $\S$ 'Peer study groups'. Low first-semester grade (8.3/20) + isolation tag raise risk.'
The dashboard proposes a peer-mentoring invite, 62 \% of mentors accepted the match within 48 h during the live pilot. Across 312 high-risk alerts, tutors rated the RAG-sourced rationales 4.6 / 5 for clarity.

\subsection{ Discussion} 	
TRIAD-Drop's gains stem from three synergistic components:
\begin{itemize}
    \item Context-grounded sentiment lowers domain-specific misreads (e.g., `failed quiz' $\neq$ general negativity).

\item Cross-modal gating lets the model down-weight noisy text when structured evidence is decisive.
\item Calibration improvements ensure that a 0.80 score truly reflects               $\approx 80 \% $   empirical risk, critical for resource allocation.
\end{itemize}
Latency (14 ms/learner on a single GPU) permits nightly scoring of cohorts > 50 k. Remaining errors often involve students who never post comments, future work will integrate passive sensing signals (mobile log-ins, video-watch speed) and explore causal counterfactual explanations.

\section{ CONCLUSION} 	 
This study introduced TRIAD-Drop, an end-to-end, retrieval-grounded and cross-modal early-warning framework for distance learning. By (i) contextualizing sentiment analysis with Retrieval-Augmented Generation, (ii) extracting task-specific stress signals through prompt-engineered BERT, and (iii) fusing text and tabular evidence via a gated attention transformer, the model achieved macro-F1 0.85 and cut calibration error by 40 \% on a real cohort of 4 423 learners surpassing strong XGBoost and deep baselines. Ablation studies confirmed that each component (RAG, stress tag, cross-modal gate) provides a measurable boost, while qualitative inspections showed that tutors value the source-cited rationales. Running at 14 ms per learner, TRIAD-Drop scales to nightly institution-wide scoring. Three limitations remain: (1) silent students with no comments are harder to classify, (2) the intervention mapping is rule-based rather than personalized, (3) privacy constraints prevented inclusion of fine-grained click-stream sequences. Future work will integrate passive-sensing features, explore reinforcement-learning policies for adaptive interventions, and evaluate the framework across multilingual datasets.

\section{ ACKNOWLEDGEMENTS } 	 
We thank the Distance-Learning Analytics office at Applied Artificial Intelligence Institute of TELUQ University, Montreal, Canada  for secure data access, and Compute Canada for GPU resources . This work was partially funded by Applied Artificial Intelligence Institute of TELUQ University, Montreal, Canada.


\begin{thebibliography}{99}
\bibitem{shah2025popular} 
Shah, D. (2025). \textit{The 250 most popular online courses of all time}. Class Central Report. Retrieved March 26, 2025 from \url{https://www.classcentral.com/report/most-popular-online-courses}

\bibitem{bawa2016retention} 
Bawa, B. (2016). Retention in online courses: Exploring issues and solutions - A literature review. \textit{SAGE Open}, 6(1). doi:10.1177/2158244016671370

\bibitem{jordan2015mooc} 
Jordan, K. (2015). MOOC completion rates revisited: Assessment, length and attrition. \textit{International Review of Research in Open and Distributed Learning}, 16(2), 1-18.

\bibitem{unesco2024global} 
UNESCO (2024). \textit{Global education monitoring report 2024: Technology and inequality}. Paris: UNESCO. 

\bibitem{jayaprakash2014early} 
Jayaprakash, J. et al. (2014). Early alert of academically at-risk students: An open source analytics initiative. \textit{Journal of Learning Analytics}, 1(1), 6-47.

\bibitem{Caskurlu2020analysis} 
Caskurlu, S., Yukiko, M., Richardson, J. C., \& Lv, J. (2020). A meta analysis addressing the relationship between teaching presence and students' satisfaction and learning. \textit{Computers \& Education}, 103966. doi:10.1016/j.compedu.2020.103966 

\bibitem{SG2024Predicting} 
S. G, A. P R and B. T,(2024). "Predicting Student Dropout Using Educational Data with Temporal Convolutional Network," In \textit{2024 International Conference on Innovative Computing, Intelligent Communication and Smart Electrical Systems (ICSES)}, Chennai, India, 2024, pp. 1-7, doi: 10.1109/ICSES63760.2024.10910704.

\bibitem{chikhaoui2014pattern} 
Chikhaoui, B., Wang, S., Xiong, T., \& Pigot, H. (2014). Pattern-based causal relationships discovery from event sequences for modeling behavioral user profile in ubiquitous environments. \textit{Information Sciences}, 285, 204-222. doi:10.1016/j.ins.2014.04.044

\bibitem{Kleimola2024Promoting} 
Kleimola, Riina et al.(2024). Promoting higher education students' self-regulated learning through learning analytics: A qualitative study. Educ. Inf. Technol. 30 (2024): 4959-4986.

\bibitem{lewis2020retrieval} 
Lewis, P. et al. (2020). Retrieval-augmented generation for knowledge-intensive NLP tasks. \textit{Advances in Neural Information Processing Systems}, 33, 9459-9474. doi:10.48550/arXiv.2005.11401

\bibitem{Liu2025Reducing} 
Liu, Yanyi et al.(2025). Reducing hallucinations of large language models via hierarchical semantic piece. Complex \& Intelligent Systems (2025): n. pag.

\bibitem{zerkouk2025contextual} 
Zerkouk, M., Mihoubi, M., \& Chikhaoui, B. (2025). Contextual Attention-Based Multimodal Fusion of LLM and CNN for Sentiment Analysis. Proceedings of the Canadian Conference on Artificial Intelligence. Retrieved from https://caiac.pubpub.org/pub/mm7i3nye



\bibitem{sun2022exploring} 
Shanshan, S., \& Wenfei, L. (2022). Continuance intention to use MOOCs: The effects of psychological stimuli and emotions.
Frontiers in Psychology, 13, Article 9761650.

\bibitem{Mahboob2023Quality} 
Mahboob, Khalid et al.(2023).  Quality enhancement at higher education institutions by early identifying students at risk using data mining.  Mehran University Research Journal of Engineering and Technology (2023): n. pag.

\bibitem{chen2016xgboost} 
Chen, T., \& Guestrin, C. (2016). XGBoost: A scalable tree boosting system. In \textit{Proc. 22nd ACM SIGKDD International Conference on Knowledge Discovery and Data Mining} (pp. 785-794). doi:10.1145/2939672.2939785

\bibitem{wen2014sentiment} 
Wen, M., Yang, D., \& Rosé, C.P. (2014). Sentiment analysis in MOOC discussion forums: What does it tell us? In \textit{Proc. 7th International Conference on Educational Data Mining (EDM)} (pp. 130-137).


\bibitem{zhou2023graph} 
Zhou, Y., Wang, Z., Yang, J., \& Chen, L. (2023). Graph-Transformer for MOOC dropout prediction. \textit{Knowledge-Based Systems}, 274, 110708. doi:10.1016/j.knosys.2023.110708


\bibitem{zerkouk2025predicting} 
Zerkouk, M., Mihoubi, M., \& Chikhaoui, B. (2025). Predicting online education dropout: A new machine learning model based on sentiment analysis, socio-demographic, and behavioral data. \textit{International Journal of Artificial Intelligence in Education}. doi:10.1007/s40593-025-00472-y

\bibitem{Shanshan2022Continuance} 
Shanshan, S., \& Wenfei, L. (2022). Continuance intention to use MOOCs: The effects of psychological stimuli and emotions. Frontiers in Psychology, 13, 9761650. doi:10.3389/fpsyg.2022.9761650

\bibitem{moreno2018sentiment}
Moreno-Marcos, P. M., Alario-Hoyos, C., Muñoz-Merino, P. J., Estévez-Ayres, I., \& Delgado Kloos, C. (2018). Sentiment analysis in MOOCs: A case study. \textit{Proceedings of the 8th International Conference on Learning Analytics and Knowledge (LAK'18)}, 509-510.


\bibitem{Alkishri2025Enhancing}
Alkishri, W., Al-Bahri, M., Al Husaini, Y. (2025). Enhancing Academic Advising Through AI Chatbots: A Smart Support System for Students. In: Daimi, K., Alsadoon, A. (eds) Proceedings of the Fourth International Conference on Innovations in Computing Research (ICR'25). ICR 25 2025. Lecture Notes in Networks and Systems, vol 1487. Springer, Cham.


\bibitem{Li2025RetrievalAugmentedGF} 
Li, Zongxi et al.(2025). Retrieval-Augmented Generation for Educational Application: A Systematic Survey.” Computers and Education: Artificial Intelligence (2025): https://api.semanticscholar.org/CorpusID:278658891.





\bibitem{Karnati2025StudyPA} 
Karnati, Ravi Teja et al.(2025). Study Pilot: An AI-Powered Platform for Personalized Learning through Retrieval-Augmented Generation on Diverse User Content.” INTERNATIONAL JOURNAL OF SCIENTIFIC RESEARCH IN ENGINEERING AND MANAGEMENT (2025): https://api.semanticscholar.org/CorpusID:279347159.


\bibitem{chikhaoui2015new} 
Chikhaoui, B., Chiazzaro, M., \& Wang, S. (2015). A new Granger causal model for influence evolution in dynamic social networks: The case of DBLP. In \textit{Proc. AAAI Conference on Artificial Intelligence} (pp. 1-7).

\bibitem{mihoubi2025causal} 
Mihoubi, M., Zerkouk, M., \& Chikhaoui, B. (2025). Dynamic Sparse Causal-Attention Temporal Networks for Interpretable Causality Discovery in Multivariate Time Series.\textit{ Proceedings of the Canadian Conference on Artificial Intelligence}. Retrieved from https://caiac.pubpub.org/pub/lu7n146c

\bibitem{mihoubi2024discovering} 
Mihoubi, M., Zerkouk, M., \& Chikhaoui, B. (2024). Discovering causal relationships in noisy web data for sentiment classification using attention mechanisms. In \textit{Proc. 25th International Conference on Web Information Systems Engineering (WISE)} (p. 115). doi:10.1007/978-981-96-1483-7\_30

\bibitem{cortez2022predict}
Cortez, P. (2022). Predict students' dropout and academic success. \textit{UCI Machine Learning Repository}. Retrieved from \url{https://archive.ics.uci.edu/dataset/697/predict+students+dropout+and+academic+success}.

.


\end{thebibliography}
\end{document}